# SQG-Differential Evolution for difficult optimization problems under a tight function evaluation budget

Ramses Sala[1,2], Niccolò Baldanzini[1], Marco Pierini[1]

[1]Department of Industrial Engineering, University of Florence, Italy
[2]rsala.unifi@gmail.com

**Abstract.** In the context of industrial engineering, it is important to integrate efficient computational optimization methods in the product development process. Some of the most challenging simulation-based engineering design optimization problems are characterized by: a large number of design variables, the absence of analytical gradients, highly non-linear objectives and a limited function evaluation budget. Although a huge variety of different optimization algorithms is available, the development and selection of efficient algorithms for problems with these industrial relevant characteristics, remains a challenge. In this communication, a hybrid variant of Differential Evolution (DE) is introduced which combines aspects of Stochastic Quasi-Gradient (SQG) methods within the framework of DE, in order to improve optimization efficiency on problems with the previously mentioned characteristics. The performance of the resulting derivative-free algorithm is compared with other state-of-the-art DE variants on 25 commonly used benchmark functions, under tight function evaluation budget constraints of 1000 evaluations. The experimental results indicate that the new algorithm performs excellent on the "difficult" (high dimensional, multi-modal, inseparable) test functions. The operations used in the proposed mutation scheme, are computationally inexpensive, and can be easily implemented in existing differential evolution variants or other population-based optimization algorithms by a few lines of program code as an non-invasive optional setting. Besides the applicability of the presented algorithm by itself, the described concepts can serve as a useful and interesting addition to the algorithmic operators in the frameworks of heuristics and evolutionary optimization and computing.

## 1 Introduction

The combination of computational optimization with modeling and simulation is becoming increasingly important in the modern development processes of complex engineering products and systems. During the last decades, a huge variety of heuristic and meta-heuristic search techniques have been developed [1, 2] and applied to real-world industrial problems [3, 4]. In the quest for product and process efficiency, an important question is: How to select efficient optimization methods for a particular problem? The extension of the conservation law of generalization of performance [5], and the "no free lunch" (NFL) theorems for machine learning to the field of search and optimization [6], identified that often "generalization is a zero sum enterprise" [5], and that therefore: the search for a universal best performing optimization algorithm is futile. The still standing challenge is: to develop and identify efficient optimization algorithms for particular optimization problems, or classes of optimization problems, taking into account the available resources in the context of their application.

In this communication, we target optimization problems under strict function evaluation budget constraints, which often occur in the context of industrial optimization problems which involve the simulation responses of complex dynamic systems. Industrial applications of such problems are for example: simulation based crashworthiness optimization of vehicle structures [9, 10, 11], and Computational Fluid Dynamics (CFD) based optimization [7, 8]. The optimization problems of such complex system responses are often characterized by: a large number of design variables, the absence of analytical gradient information, highly non-linear system responses, and computationally expensive function evaluations resulting in a limited function evaluation budget. The need to adapt the engineering development and optimization process to products and systems with increasing complexity make the research for efficient optimization algorithms, for non-convex optimization problems, under tight function evaluation constraints of great relevance in engineering [3, 4, 10, 11, 12, 13].

Despite recent and ongoing research on the theoretical performance analysis of heuristic search algorithms on fixed budget problems [14], the performance analysis of complex problems and optimization algorithm is in practice still restricted to numerical comparative tests. The optimization algorithm performance comparisons in the literature are however often w.r.t. algorithm convergence behavior using a large number (hundreds of thousands to millions) of function evaluations. For engineering optimization problems which involve computationally expensive simulations, the function evaluation budget is often orders of magnitudes smaller, such that true optimization near to the global optimum is often infeasible [10, 24]. When the function evaluation budget strongly constraints the optimization, different aspects of the optimization algorithms are of practical relevance.

In this communication, we present a new variant of the well-known and widely used Differential Evolution (DE) algorithm [15, 16], by introducing a novel mutation operator inspired by concepts of Stochastic Quasi-Gradient (SQG) methods [25, 26]. The new hybrid algorithm targets to improve the search efficiency in the setting of optimization under tight function evaluation budget constraints. To investigate the effect of the new DE mutation operator, the performance of the hybrid algorithms is compared with "classical" DE and several state-of-the-art DE variants [17, 18, 19, 20, 21], on a commonly used set of test functions of various structure and complexity, under budget constraints of 1000 function evaluations. Although the mutation operator could also be used in other similar algorithm classes such as Particle Swarm Optimization (PSO) [29], or Evolutionary Strategies (ES), the focus of this first study will be limited to the implementation and performance comparison of the SQG-mutation operator in the framework of Differential Evolution and several of its state-of-the-art variants. In the context of budget limited optimization problems in structural and multidisciplinary optimization, DE was identified and recommended as an efficient algorithm for car-body optimization problems involving computationally expensive crashworthiness responses [9, 10, 11, 12]. DE algorithms are also used for optimization of aircraft engines [7], wind turbines [8], and many other applications [22, 23]. Optimization problems in the "expensive" function evaluation setting can also benefit from meta-modelling or surrogate model based optimization techniques such as e.g. [27, 28]. In-depth investigations on the interactions between optimization algorithm operators and different meta-models and control parameters require however an extended scope. Nevertheless, even in the present limited scope, the obtained results indicate that the new algorithm, could already be an efficient alternative to several state-of-the-art DE variants, for difficult problems under a tight function evaluation budget.

## 2 Description of the SQG-DE algorithm

### 2.1 Conceptual description

The objective in the design of heuristic and meta-heuristic optimization algorithms is to obtain a beneficial compromise between efficiency and accuracy (or optimality), taking into account the available resources. The here presented hybrid method is developed to improve the efficiency of DE for a variety of problem types under strict evaluation budgeted constraints. This is achieved by means of a new mutation operator. While in conventional DE, the mutation operator uses a sum of random vector differences from the DE population, in the new mutation operator the perturbation directions of the new population members are constructed by a weighted sum approach, using the weights dependent of respective fitness differences. This concept for the perturbation directions was inspired by the stochastic quasi-gradient estimations [25, 26] used in the SQG method. Whereas in SQG-descend the stochastic gradient estimations are based on vector differences of small stochastic perturbations, the here described method applies the concept to vector differences of the DE population.

### 2.2 Quantitative description

The new hybrid SQG-DE algorithm uses the framework of the conventional original Differential Evolution (DE) algorithm. For the description of the relatively well-known DE algorithm we refer to [15, 16], while for an overview of variants we refer to the reviews in [22, 23]. For the here proposed hybrid method a new mutation operator was developed, which will be described in this section. The new mutation operator was inspired by the Stochastic Quasi-Gradient (SQG) method (initially introduced as: "search by means of statistical gradients"). A detailed description of stochastic quasi-gradient methods is given in [26]. For the sake of clarity and briefness, the description here is limited to concepts relevant for the new mutation operator.
In SQG, the search direction is the stochastic gradient approximation $\xi(x)$ of a function $f(x)$ at point $x$. This direction is proportional to the following expression:

$$\xi(x) \sim \sum_{k=1}^{r} \left( \frac{(f(x+\Delta z_k) - f(x))}{\Delta} \right) * z_k \quad (1)$$

Where $z_k \in [-1,1]^D$ are uniform random perturbation vectors of current trial vector $x$ in dimension D. For a sufficiently large $r$, and sufficiently small values of $\Delta$ this approximation converges in probability to the direction of the gradient $\nabla f$. SQG is however often applied using $r$ significantly smaller than the problem dimension D, leading to coarse and "inexpensive" gradient approximations. Compared to other gradient based methods that require finite difference gradient approximations, SQG often however performs surprisingly well on local search problems, considering the efficiency in terms of the total amount of function evaluations. The idea of approximating the gradient direction $d$ at point $x$, by means finite perturbations of the trial vector $x$, can be generalized to a sum of differences between pairs of distinct vectors $x_a$ and $x_b$, in a sufficiently small neighborhood $\varepsilon$ of $x$ with: $\|x - x_a\| < \varepsilon$, and $\|x - x_b\| < \varepsilon$ by:

$$d \sim \frac{1}{w} \sum_{k=1}^{w} \frac{(f(x_{a,k}) - f(x_{b,k}))}{\|x_{a,k} - x_{b,k}\|} * (x_{a,k} - x_{b,k}) \quad (2)$$

The key concept of the proposed hybrid DE method is the extension of this concept for gradient estimation to the application of finding new mutation vectors based on an existing differential evolution population. This extension thus omits the neighborhood constraint on a sufficiently small $\varepsilon$, and uses the differences between members of the population at a given iteration of a population based algorithm.

The mutation operator for the originally proposed "DE/rand/1/bin" version of DE [15] is determined by:

$$\boldsymbol{v_i} = \boldsymbol{x_a} + F(\boldsymbol{x_b} - \boldsymbol{x_c}) \qquad (3)$$

where $\boldsymbol{v_i}$ are the mutant vectors for the next generation, $\boldsymbol{x_q}$ are parent vectors of the current population generation, with mutually exclusive ($a \neq b \neq \cdots \neq q$) random permutation indices $a, b, \ldots, q \in \{1, 2, \ldots, P\}$ to population members, in a population of size $P$. The scaling factor $F \in [0,2]$ controls the amplification or step size of the differential variation. Later DE versions were introduced [16] such as "DE/best/2/bin", in which the mutation operator was based on a sum of more vector differences:

$$\boldsymbol{v_i} = \boldsymbol{x_{best}} + F((\boldsymbol{x_a} - \boldsymbol{x_b}) + (\boldsymbol{x_c} - \boldsymbol{x_d})) \qquad (4)$$

In which $\boldsymbol{x_{best}}$ is the best member of the population as opposed to a random member as in (3).

Combining the previous considerations, we introduce the SQG-DE hybrid scheme "SQG-DE/best/w/bin" with the SQG-mutation operator defined as:

$$\boldsymbol{v_i} = \boldsymbol{x_{best}} - F * \varphi * \sum_{k=1}^{w} \frac{(y_{b,k} - y_{c,k})}{\|x_{b,k} - x_{c,k}\|} (\boldsymbol{x_{b,k}} - \boldsymbol{x_{c,k}}) \qquad (5)$$

where $y_{q,k}$ refers to the fitness or function evaluation value corresponding to the parent vector $\boldsymbol{x_{q,k}}$ as $y_{q,k} = f(\boldsymbol{x_{q,k}})$, and $w$ is the number of mutually exclusive vector pairs used. To preserve the population self-adaptivity of the DE algorithm, a scaling factor $\varphi$ on the perturbation magnitude is included. This factor is chosen as:

$$\varphi = \frac{1/w \, \|\sum_{k=1}^{w}(x_{b,k} - x_{c,k})\|}{\left\|\sum_{k=1}^{w} \frac{(y_{b,k} - y_{c,k})}{\|x_{b,k} - x_{c,k}\|}(x_{b,k} - x_{c,k})\right\|} \qquad (6)$$

In which the denominator normalizes the magnitude of the perturbation direction, while the numerator scales the perturbation magnitude to a similar magnitude as the mutation operator in the original algorithm (3). The mutation formulation in equation (5) is similar to the original concept of vector differences, with the difference that now a sum of weighted vector differences is used, with a particular choice for the weights. The weights for the vector differences are calculated according to the fitness differences between the corresponding population members, such that in a high-dimensional setting, directions with larger directional "differences" are prioritized over directions for which the fitness differences are smaller.

Fitness differences are also used implicitly in the context of PSO, where single point pair differences between the global and local best-known locations are used to "guide" the search directions resulting from the mutation operator. In contrast, the basic SQG-mutation operator uses fitness difference based weighed sums of point pair differences between any population members, to guide the randomization of the search points. While the SQG concept is originally aimed at local gradient approximation, this is not the primary aim of the SQG-mutation operator. The extension of the SQG concept from perturbation points in a local neighborhood to random mutually exclusive population members, is likely to result in inaccurate local gradient approximations since the distance between the population members can be relatively large, and the fitness functions are generally non-linear. Since in "classical" DE the mutation operator is however only intended to introduce randomization of the population in its original implementation, there is no mechanism to favor any particular search direction. New non-descending SQG-mutation search points based on inaccurate gradient estimates are 5therefore also not problematic in the context of DE. However, for problems in which there are global trend directions, the SQG-mutations are statistically biased towards global descend directions. The key idea of the new method is however that SQG-mutations, tend to favor search directions along which the fitness differences are larger, over directions with small differences. This "direction-screening" is particularly relevant when, not all problem dimensions are equally important. For the mutation operator, parameter values of $w$ between 2 and 5 give very satisfactory results, based on our current experience. It should be noted that for very small population sizes, high values of $w$ should be avoided to maintain sufficient variance in future populations.

Although eqn. (5) is more complex than eqn. (3) only a relatively modest number of scalar and vector operations is required for the new mutation operator, while no additional function evaluations are required. The mutation operator can be easily implemented as an optional setting in existing code of DE or other population-based optimization algorithm implementations. On request, a MATLAB/Octave or other implementation of the algorithm is available from the authors.

## 3 A comparison of algorithm performance on a constrained function evaluation budget

To investigate the performance of the new hybrid algorithm, a comparative assessment of SQG-DE, with two of the original DE algorithms, the SQG algorithm, and 5 state-of-the-art DE variants is performed, on a set of 25 test functions, using a budget of maximum 1000 function evaluations.

### 3.1 Algorithms and Test functions

For the comparative assessment the following optimization algorithms are used:

1. DE       -original Differential Evolution "rand/1/exp"                    [15]
2. DE2      -"best/2/bin" Differential Evolution                              [16]
3. jDE      -self-adapting Differential Evolution                             [17]
4. JADE     -adaptive Differential Evolution                                  [20]
5. SaDE     -Strategy adaptation Differential Evolution                       [18]
6. epsDE    -ensemble parameters Differential Evolution                       [19]
7. CoDe     -Composite trial vector strategy Differential Evolution           [21]
8. SQG      -Stochastic Quasi-Gradient search                                 [25,26]
9. SQG-DE   -Stochastic Quasi-Gradient based Differential Evolution

To assess and compare the performance of the different algorithms the 25 test functions of the CEC 2005 benchmark [30] were used. Although many more benchmark sets have been developed since, these test functions are widely used in the optimization community (more than 1500 citations at present). The test function set is composed of optimization problems in 4 categories. One of these categories is of particular interest in the context of this work: The 4[th] category of "difficult" inseparable complex multimodal rotated functions, of which many are even hard to solve with a large function evaluation budget. Although these functions are usually used in the conventional context of global optimization, without tight function evaluation limits (typical budgets of hundreds of thousands to millions of objective function evaluations), they are also of interest as surrogate-test problems, in the context of algorithm performance assessment for complex industrial problems under tight function evaluation constraints. In this assessment the number of function evaluations per optimization run is constrained to a maximum of 1000. The test functions were evaluated for dimensions 30 and 50. Table 1 gives an overview of the test function descriptions and the problem categories. For a more detailed description of the test functions we refer to the description in [30].

For the comparison, the optimization runs of each algorithm were independently repeated with different random seeds for 100 times, for each test function to obtain statistically significant performance results. For all algorithms except SQG, the initial population size was set to 100, for all problems and dimension as was also done in previous works [31, 32]. For SQG a "warm" start was provided by choosing the best start point from a pseudo-random sample set of equal size as the population size of the other algorithms. The control parameters for DE, DE2 were F=0.8, CR=0.8, and in addition w=5 for and SQG-DE, and SQG. For the five other algorithms with adaptive parameters, the control parameters were as described in the corresponding references [17-21]. For the adaptive parameter algorithms, the implementations as available in [34] were used.

**Table 1.** List of test functions

| Function nr. | Test function description from [30] |
|---|---|
| **1 Unimodal Functions (5):** | |
| $f_1(x)$ | Shifted Sphere Function |
| $f_2(x)$ | Shifted Schwefel's Problem 1.2 |
| $f_3(x)$ | Shifted Rotated High Conditioned Elliptic Function |
| $f_4(x)$ | Shifted Schwefel's Problem 1.2[1] |
| $f_5(x)$ | Schwefel's Problem 2.6[2] |
| **2 Multimodal Basic Functions (7):** | |
| $f_6(x)$ | Shifted Rosenbrock's Function |
| $f_7(x)$ | Shifted Rotated Griewank's Function* |
| $f_8(x)$ | Shifted Rotated Ackley's Function[2] |
| $f_9(x)$ | Shifted Rastrigin's Function |
| $f_{10}(x)$ | Shifted Rotated Rastrigin's Function |
| $f_{11}(x)$ | Shifted Rotated Weierstrass Function |
| $f_{12}(x)$ | Schwefel's Problem 2.13 |
| **3 Multimodal Expanded Functions (2):** | |
| $f_{13}(x)$ | Expanded Extended Griewank's plus Rosenbrock's Function |
| $f_{14}(x)$ | Shifted Rotated Expanded Scaffer's F6 |
| **4 Multimodal Hybrid Composition Functions (11):** | |
| $f_{15}(x)$ | Hybrid Composition Function |
| $f_{16,18,21,24}(x)$ | Rotated Hybrid Composition Functions |
| $f_{17}(x)$ | Rotated Hybrid Composition Function[1] |
| $f_{20}(x)$ | Rotated Hybrid Composition Function[3] |
| $f_{19}(x)$ | Rotated Hybrid Composition Function[2] |
| $f_{22}(x)$ | Rotated Hybrid Composition Function[4] |
| $f_{23}(x)$ | Non-Continuous Rotated Hybrid Composition Function |
| $f_{25}(x)$ | Rotated Hybrid Composition Function |

---

[1] With noise in fitness function

[2] With the global optimum on the bounds

[3] With a narrow basin for the global optimum

[4] With a high condition number matrix

### 3.2 Performance measures

A commonly used optimization algorithm performance metric is Expected Running Time (ERT) [33], which can be defined as:

$$\text{ERT}(f_{\text{target}}) = \text{mean}\left(T_{f_{\text{target}}}\right) + ((1 - p_s)/p_s)T_{max} \quad (7)$$

Where $f_{\text{target}}$ is a reference threshold value, $T_{f_{\text{target}}}$ is the number of function evaluations to reach an objective value better than $f_{\text{target}}$, $T_{max}$ is the maximum number of function evaluations per optimization run, and $p_s$ is the success rate defined as: $p_s = N_{\text{succes}}/N_{\text{total}}$, where $N_{\text{succes}}$ is the number of successful runs (where the best obtained objective value is better than $f_{\text{target}}$). If the experiments result in no successful runs for a particular algorithm such that ($p_s = 0$), then expression (7) is undefined, in that case the information available on the ERT is that: $\text{ERT}(f_{\text{target}}) > T_{max} * N_{\text{total}}$.

ERT can be interpreted as the expected number of function evaluations of an algorithm to reach an objective function threshold for the first time. For the ERT performance measure, a threshold or success criterion is required. For conventional optimization performance studies this criterion is often related to reaching the value of the known global optimum, within a specified tolerance. For the optimization of difficult problems under tight budget constraints the probability of coming close to the global optimum is usually statistically negligible, therefore an alternative success criterion is required. To compare qualitative performance using ERT it is necessary that all compared algorithms meet the success criterion at least a few times. For the optimization performance assessment under tight budget evaluation restrictions, we define the success criterion as reaching a target value which corresponds to the expected value of the best objective function value obtained from uniform random sampling with the given function evaluation budget (1000 samples). For a test function $f_k$ we will refer to the expected objective value as $E_{f_k}^{RSE}$. The estimation of $E_{f_k}^{RSE}$ is based on the same number of repetitions as is used to measure the performance of the other algorithms (100 in this case). We will refer to the ERT w.r.t. this objective function value limit as Random Sampling Equivalent-Expected Run Time (ERT$_{\text{RSE}}$).

Besides the fixed target performance measure ERT, a further (more intuitive) way to compare the performance of the optimization algorithms on the test functions is by means of diagrams on which the Best Function Value (BFV) of the objective functions, is plotted against the number of algorithm iterations or Function evaluations. For the diagrams in the results section, the BFV has been normalized (BNFV) with $E_{f_k}^{RSE}$, for the corresponding test function. The advantage of these diagrams is that they give an intuitive picture of performance for both, fixed-cost, and fixed-target scenarios.

Except for the ERT reference value, and an increased number of repetitions, the experimental set up of the study followed the benchmark description in [30]. To assess the significance of the overall performance test results, the non-parametric Wilcoxon signed rank test [36] was applied pairwise between the results of the algorithms, with the best algorithm as the reference (see also [35]). The null hypothesis of this test is: a zero difference of the median between two results sets. The conventional significance threshold of 0.05 is used to indicate that the null hypothesis cannot be rejected with sufficient certainty.

### 3.3 Results Comparison

All test problems in this comparison are minimization problems. Good optimization algorithm performance is thus related to reaching a low BNFV in few function evaluations. For the 50-dimensional problem set, BNFV diagrams comparing the algorithm performance for the 9 algorithms are displayed in **Fig. 1**. The SQG-DE algorithm ranked as the best algorithm in terms of BNFV performance after 1000 function evaluations in 16 out of the 25 test problems and was the winner in all of the 10 test problems of the 4[th] category (Multi-modal Hybrid composition functions). The results for the 30-dimensional problem set were similar, but not reported in a figure due to space constraints.

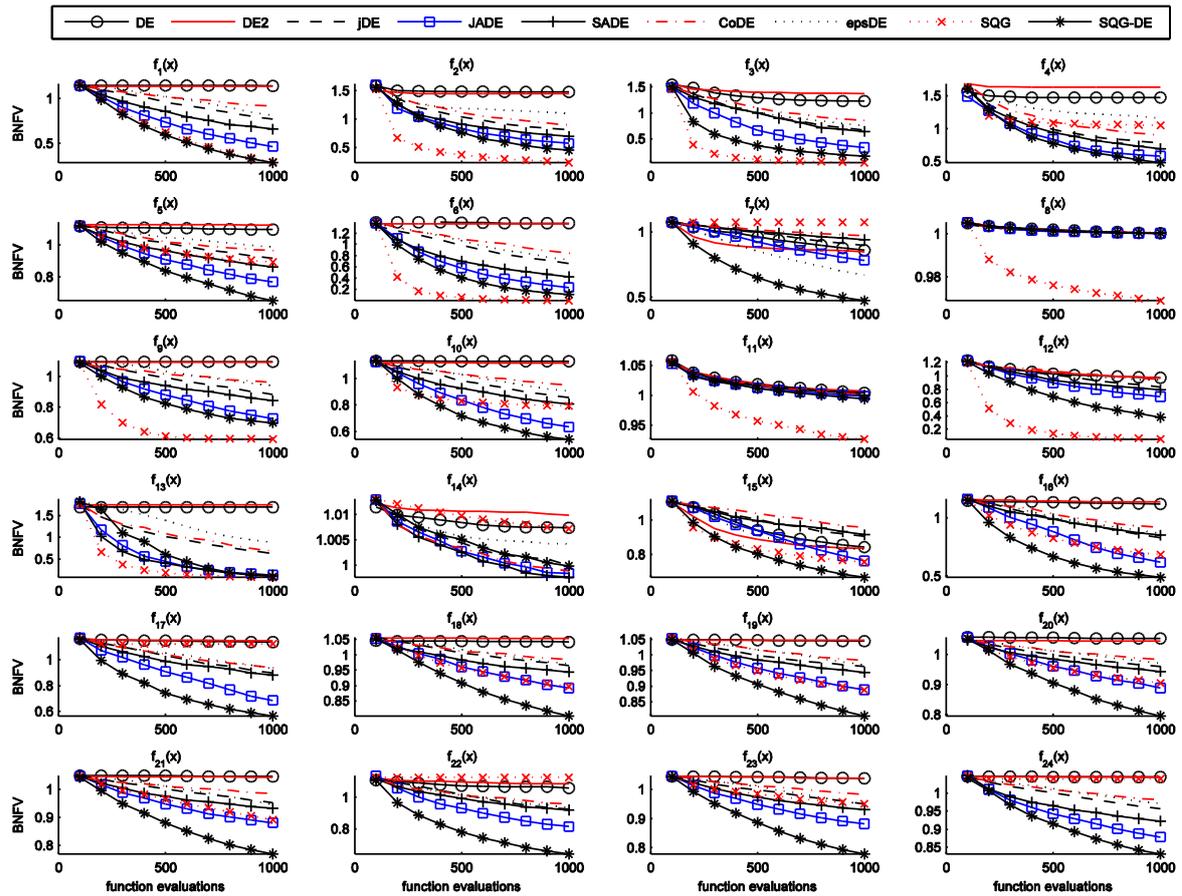

**Fig. 1.** Evolution of the Best Normalized Function Value (BNFV) for increasing function evaluations, functions 1-24 (D=50).

An overview of algorithm performance in terms of the ERT$_{\text{RSE}}$ for all 30 and 50 dimensional problems is given in **Table 2**. An ERT$_{\text{RSE}}$-value of for example 300 means that the corresponding algorithms requires 300 function evaluations to obtain a function evaluation better than the threshold, (which was defined as the expected best objective value for 1000 uniform random samples in the problem domain). **Table 2** shows that SQG-DE achieves the best ERT performance in 30 out of the 50 test problems. The new hybrid algorithm performed also with respect to the ERT measure as the best in all of the test problems of the 4[th] category.

**Table 2.** Algorithm performance in terms of ERT$_{\text{RSE}}$, for all test functions

| Test functions | Dimension 30 | | | | | | | | | Dimension 50 | | | | | | | | |
|---|---|---|---|---|---|---|---|---|---|---|---|---|---|---|---|---|---|---|
| | DE | DE2 | jDE | JADE | SADE | Code | epsDE | SQG | SQG-DE | DE | DE2 | jDE | JADE | SADE | Code | epsDE | SQG | SQG-DE |
| **1 Unimodal** | | | | | | | | | | | | | | | | | | |
| $f_1(x)$ | 11657 | 24049 | 491 | 264 | 329 | 904 | 606 | 202 | **168** | 19030 | 9054 | 439 | 205 | 254 | 674 | 579 | 192 | **178** |
| $f_2(x)$ | 7401 | 11567 | 739 | 309 | 379 | 894 | 3653 | **137** | 318 | 9055 | 8145 | 572 | 316 | 388 | 908 | 2135 | **127** | 296 |
| $f_3(x)$ | 6536 | 32763 | 606 | 356 | 519 | 1148 | 1045 | **135** | 210 | 3819 | 10317 | 514 | 289 | 455 | 764 | 566 | **121** | 145 |
| $f_4(x)$ | 6182 | 13326 | 570 | **285** | 342 | 662 | 1712 | 694 | 310 | 19059 | 10175 | 538 | 341 | 434 | 920 | 2931 | 1332 | **310** |
| $f_5(x)$ | 4342 | 4766 | 536 | 311 | 527 | 871 | 564 | 459 | **160** | 9173 | 19040 | 568 | 294 | 366 | 964 | 1418 | 451 | **214** |
| **2 Multimodal Basic** | | | | | | | | | | | | | | | | | | |
| $f_6(x)$ | 7404 | 11728 | 491 | 234 | 315 | 807 | 603 | **124** | 168 | 11557 | 13327 | 491 | 235 | 244 | 831 | 609 | **132** | 182 |
| $f_7(x)$ | 256 | 124 | 482 | 310 | 596 | 879 | 169 | 32365 | **107** | 394 | 187 | 443 | 289 | 536 | 884 | 231 | 24041 | **109** |
| $f_8(x)$ | 2877 | 3011 | 1783 | 2084 | 2673 | 2720 | 2082 | **132** | 2771 | 1732 | 2347 | 2098 | 1792 | 2724 | 2522 | 2740 | **117** | 1960 |
| $f_9(x)$ | 15735 | 24058 | 467 | 269 | 380 | 818 | 726 | **126** | 193 | 15719 | 19060 | 482 | 245 | 295 | 845 | 738 | **123** | 194 |
| $f_{10}(x)$ | 11658 | 24054 | 474 | 256 | 320 | 850 | 630 | 201 | **172** | 24071 | 24059 | 456 | 235 | 306 | 739 | 661 | **169** | 188 |
| $f_{11}(x)$ | 2414 | 1555 | 1967 | 1502 | 1321 | 1541 | 1979 | **368** | 1949 | 2024 | 2568 | 2187 | 1324 | 1814 | 1825 | 1675 | **347** | 1331 |
| $f_{12}(x)$ | 1072 | 934 | 509 | 342 | 416 | 931 | 594 | **128** | 225 | 1183 | 1263 | 560 | 340 | 418 | 1132 | 774 | **119** | 224 |
| **3 Multimodal Expanded** | | | | | | | | | | | | | | | | | | |
| $f_{13}(x)$ | 15709 | 10150 | 524 | 224 | 231 | 723 | 898 | **161** | 289 | 7381 | 9057 | 514 | 227 | 193 | 604 | 1139 | **145** | 336 |
| $f_{14}(x)$ | 7497 | 19174 | 2735 | 1735 | **1509** | 2366 | 6379 | 6899 | 1658 | 7476 | 15824 | 1792 | 1310 | **1198** | 1296 | 3482 | 4706 | 1808 |
| **4 Multimodal Hybrid** | | | | | | | | | | | | | | | | | | |
| $f_{15}(x)$ | 437 | 373 | 668 | 366 | 565 | 857 | 458 | 1016 | **185** | 341 | 291 | 557 | 354 | 556 | 852 | 331 | 205 | **173** |
| $f_{16}(x)$ | 2999 | 11671 | 490 | 335 | 615 | 689 | 616 | 219 | **179** | 7506 | 11568 | 495 | 320 | 487 | 738 | 593 | 260 | **156** |
| $f_{17}(x)$ | 6925 | 11670 | 514 | 362 | 572 | 845 | 693 | 9096 | **178** | 11585 | 32369 | 640 | 327 | 515 | 794 | 809 | 5743 | **192** |
| $f_{18}(x)$ | 1017 | 1036 | 501 | 314 | 460 | 741 | 341 | 263 | **143** | 10264 | 19135 | 631 | 319 | 380 | 1071 | 892 | 352 | **226** |
| $f_{19}(x)$ | 1271 | 1045 | 533 | 317 | 551 | 836 | 389 | 259 | **148** | 24351 | 7387 | 511 | 290 | 389 | 866 | 728 | 307 | **210** |
| $f_{20}(x)$ | 1190 | 1098 | 491 | 315 | 545 | 1039 | 384 | 284 | **154** | 19307 | 13407 | 533 | 321 | 372 | 915 | 794 | 380 | **227** |
| $f_{21}(x)$ | 9466 | 24251 | 498 | 253 | 327 | 901 | 597 | 301 | **175** | 13571 | 9089 | 486 | 243 | 320 | 871 | 605 | 325 | **176** |
| $f_{22}(x)$ | 2433 | 2889 | 489 | 281 | 389 | 730 | 602 | 13320 | **202** | 3311 | 5068 | 564 | 291 | 431 | 755 | 724 | 19048 | **153** |
| $f_{23}(x)$ | 13650 | 24194 | 474 | 254 | 325 | 802 | 586 | 664 | **181** | 7468 | 7544 | 483 | 248 | 333 | 880 | 590 | 537 | **170** |
| $f_{24}(x)$ | 7785 | 4449 | 544 | 272 | 376 | 911 | 573 | 10183 | **187** | 15709 | 13335 | 498 | 231 | 245 | 784 | 688 | 8170 | **201** |
| $f_{25}(x)$ | 220 | 115 | 445 | 304 | 554 | 855 | 167 | 5321 | **109** | 304 | 163 | 488 | 315 | 552 | 1038 | 210 | 13368 | **109** |

Table 3 provides a summary of algorithm performance in terms of $ERT_{RSE}$, divided by test function category, and averaged overall performance. The results in Table 3 indicate that: SQG-DE had the best average performance over all test problems, with an ERT of approximately 10% less w.r.t. the second overall best algorithm JADE. More remarkable is that in the 4th category with the hardest test functions, SQG-DE obtained ERT scores which are about 40% better than the second-best algorithm (JADE). The Wilcoxon signed rank test indicated that the all the results were statistically significant, except for the small test problem group 3. Closer inspection revealed that this was exclusively caused by test function 14, for which all of the investigated algorithms performed worse than random sampling, which indicates that $f_{target}$ was rarely reached.

The results from this benchmark indicate that for hard high dimensional multimodal, problems under a tight function evaluation budget the new hybrid algorithm performs significantly better than the original "parent" algorithms DE, SQG, and the state-of-the-art DE variants tested.

**Table 3.** Overview of algorithm performance in terms of $ERT_{RSE}$, by test function category

| Test function groups | Dimension | DE | DE2 | jDE | JADE | SADE | Code | epsDE | SQG | SQG-DE | p 1st. rank |
|---|---|---|---|---|---|---|---|---|---|---|---|
| **1 Unimodal** | D=30 | 7223 | 17294 | 588 | 305 | 419 | 896 | 1516 | 325 | **233** | p<0.01 |
| $f_{1-5}$ | D=50 | 12027 | 11346 | 526 | 289 | 379 | 846 | 1526 | 444 | **228** | p<0.01 |
| **2 Multimodal Basic** | D=30 | 5917 | 9352 | 882 | **714** | 860 | 1221 | 969 | 4778 | 798 | p<0.01 |
| $f_{6-12}$ | D=50 | 8097 | 8973 | 960 | 637 | 905 | 1254 | 1061 | 3578 | **598** | p<0.01 |
| **3 Multimodal Expanded** | D=30 | 11603 | 14662 | 1630 | 980 | **870** | 1544 | 3638 | 3530 | 974 | p=0.197 |
| $f_{13-14}$ | D=50 | 7429 | 12440 | 1153 | 768 | **696** | 950 | 2311 | 2426 | 1072 | p=0.014 |
| **4 Multimodal Hybrid** | D=30 | 4309 | 7526 | 513 | 307 | 480 | 837 | 491 | 3721 | **167** | p<0.01 |
| $f_{15-25}$ | D=50 | 10338 | 10851 | 535 | 296 | 416 | 869 | 633 | 4427 | **181** | p<0.01 |
| **Overal performance** | D=30 | 5925 | 10562 | 721 | 474 | 605 | 1013 | 1082 | 3322 | **422** | p<0.01 |
| $f_{1-25}$ | D=50 | 9816 | 10551 | 702 | 428 | 568 | 979 | 1066 | 3233 | **379** | p<0.01 |

## 4 Discussion and outlook

The performance comparison results show efficiency gains of SQG-DE ranging up to 40%, w.r.t. the next best algorithm in the category of Multimodal Hybrid Composition functions. Overall the performance benefits of SQG-DE w.r.t. the "parent" algorithms (SQG and DE2 "best/2/bin") indicates a useful synergy effect, which already could be exploited to solve complex budget constrained optimization problems, in its present state.

The remarkable results also call for further activities and investigations, such as: further performance comparisons against optimization algorithms other than DE; implementing the SQG-mutation operator in other DE variants; control parameter tuning; implementation of suitable self-adaptive parameters strategies; and hybridization of SQG with other population-based meta-heuristic algorithms such as ES and PSO.

In the present study the conventional control parameters settings, according to the recommendations in the respective literature, were used for the optimization algorithms. The best choice for the control parameters, is however both problem and budget dependent. Budget dependent control parameter tuning or optimization, for a particular test function or set of test functions is possible. For computationally expensive industrial optimization problems such parameter tuning is however several orders of magnitude more expensive than an optimization run, such that direct control parameter optimization on real-world problems is often infeasible. Although it is possible to optimize or tune the algorithms settings on conventional synthetic test or benchmark problems, it is important in the context of industrially relevant problems to know or estimate the algorithm performance correlations between the synthetic test problems, and a given real-world problem. The industrial relevance of detailed comparative studies including new algorithms, operators or tuned control parameters, are relative to the quantifiability of performance correlations with real-world problems, or by the gained theoretical insights. We are however obliged to note that performance on most of the conventional synthetic benchmark functions (including those used for this study) is difficult to relate (or quantitatively correlate) to performance on particular real-world optimization problems, which is thus a strong limiting factor for direct practical relevance. Also the theoretical insights and generalizability of the results are limited by the lack of systematic relations among the conventional test functions. These strong limitations apply to the presented study, as well as to most of the work in the literature which is based on conventional synthetic test problems and benchmark sets.

In order to obtain systematic results that could lead to insights of theoretical value, and improved optimization performance in real-world problems, most of the earlier mentioned plans for further investigations on the SQG-mutation operator will be performed using test functions with parameterized function characteristics such as presented in [37], benchmarks based on engineering design optimization applications, and new synthetic test approaches such as representative surrogate problems [12]. Important open questions are: How are the performance of SQG-DE and other meta-heuristic optimization algorithms related to particular problem characteristics? How do the control parameters interact with problem characteristics in terms of algorithm performance? Further investigations and insights are required to address these questions.

## 5 Conclusions

A new SQG inspired mutation operator is introduced in the framework of DE, resulting in a new hybrid algorithm "SQG-DE". The algorithm is compared with conventional DE and several state-of-the-art DE variants, w.r.t. optimization performance under strict function evaluation budget constraints. The results of the comparison indicate that the new algorithm excels the other compared algorithms on average by 10% in overall performance, on the investigated benchmark problems. The new algorithm performs particularly well on high dimensional multi-modal composite test problems (of the $4^{th}$ test problem category), where w.r.t. fixed target performance measures, averaged function evaluation savings of about 40% are achieved. The results are promising, and the displayed optimization efficiency could be of relevance for Industrial real-world problem settings, which involve a strict function evaluation budget. The described mutation operator is computationally inexpensive, easy to implement and could therefore also be used in other population-based meta-heuristic optimization approaches. On request, an implementation of the SQG-DE algorithm is available from the authors.

## 6    Acknowledgments

This work was partially funded by the GRESIMO project grant agreement no. 290050 by the European community 7[th] Framework program. We would like to thank the anonymous reviewers for their remarks to improve the manuscript. Furthermore, we like to express our gratitude to Qingfu Zhang and all other cited authors who made code of their algorithms and test benches publicly available on their websites.